\documentclass[conference]{IEEEtran}
\IEEEoverridecommandlockouts
\usepackage{cite}
\usepackage{amsmath,amssymb,amsfonts}
\usepackage{algorithmic}
\usepackage{graphicx}
\usepackage{textcomp}
\usepackage{xcolor}
\usepackage{algorithm}
\usepackage{algorithmic}
\usepackage{amsmath}
\usepackage{makecell}
\usepackage{hyperref}
\def\BibTeX{{\rm B\kern-.05em{\sc i\kern-.025em b}\kern-.08em
    T\kern-.1667em\lower.7ex\hbox{E}\kern-.125emX}}
\DeclareRobustCommand*{\IEEEauthorrefmark}[1]{%
    \raisebox{0pt}[0pt][0pt]{\textsuperscript{\footnotesize\ensuremath{#1}}}}

\begin{document}

\title{TSTMotion: Training-free Scene-aware Text-to-motion Generation}

\makeatletter
\newcommand{\linebreakand}{%
  \end{@IEEEauthorhalign}
  \hfill\mbox{}\par
  \mbox{}\hfill\begin{@IEEEauthorhalign}
}
\makeatother










\author{
\thanks{* These authors contributed equally and † is corresponding author.}
\IEEEauthorblockN{
1\textsuperscript{st} Ziyan Guo\IEEEauthorrefmark{1}$^{,*}$,
2\textsuperscript{nd} Haoxuan Qu\IEEEauthorrefmark{2}$^{,*}$,
3\textsuperscript{rd} Hossein Rahmani\IEEEauthorrefmark{2}, 
4\textsuperscript{th} Dewen Soh\IEEEauthorrefmark{2},  \\
5\textsuperscript{th} Ping Hu\IEEEauthorrefmark{3}, 
6\textsuperscript{th} Qiuhong Ke\IEEEauthorrefmark{4},
and 7\textsuperscript{th} Jun Liu\IEEEauthorrefmark{2}$^{,\dag}$}
\IEEEauthorblockA{\IEEEauthorrefmark{1}Singapore University of Technology and Design, Singapore}
\IEEEauthorblockA{\IEEEauthorrefmark{2}Lancaster University, Lancaster, England}
\IEEEauthorblockA{\IEEEauthorrefmark{3}University of Electronic Science and Technology of China, Chengdu, China}
\IEEEauthorblockA{\IEEEauthorrefmark{4}Monash University, Victoria, Australia}
\IEEEauthorblockA{ziyan\_guo@mymail.sutd.edu.sg, \{h.qu5, h.rahmani, j.liu81\}@lancaster.ac.uk, \\ dewen\_soh@sutd.edu.sg, chinahuping@gmail.com, Qiuhong.Ke@monash.edu, }
}

\maketitle
\begin{abstract}
Text-to-motion generation has recently garnered significant research interest, primarily focusing on generating human motion sequences in blank backgrounds. However, human motions commonly occur within diverse 3D scenes, which has prompted exploration into scene-aware text-to-motion generation methods. Yet, existing scene-aware methods often rely on large-scale ground-truth motion sequences in diverse 3D scenes, which poses practical challenges due to the expensive cost. To mitigate this challenge, we are the first to propose a \textbf{T}raining-free \textbf{S}cene-aware \textbf{T}ext-to-\textbf{Motion} framework, dubbed as \textbf{TSTMotion}, that efficiently empowers pre-trained blank-background motion generators with the scene-aware capability. Specifically, conditioned on the given 3D scene and text description, we adopt foundation models together to reason, predict and validate a scene-aware motion guidance. Then, the motion guidance is incorporated into the blank-background motion generators with two modifications, resulting in scene-aware text-driven motion sequences. Extensive experiments demonstrate the efficacy and generalizability of our proposed framework. We release our code in \href{https://tstmotion.github.io/}{Project Page}.
\end{abstract}


\section{INTRODUCTION}

\begin{figure}[!t]
  \centering
  \includegraphics[width=\linewidth,page=1]{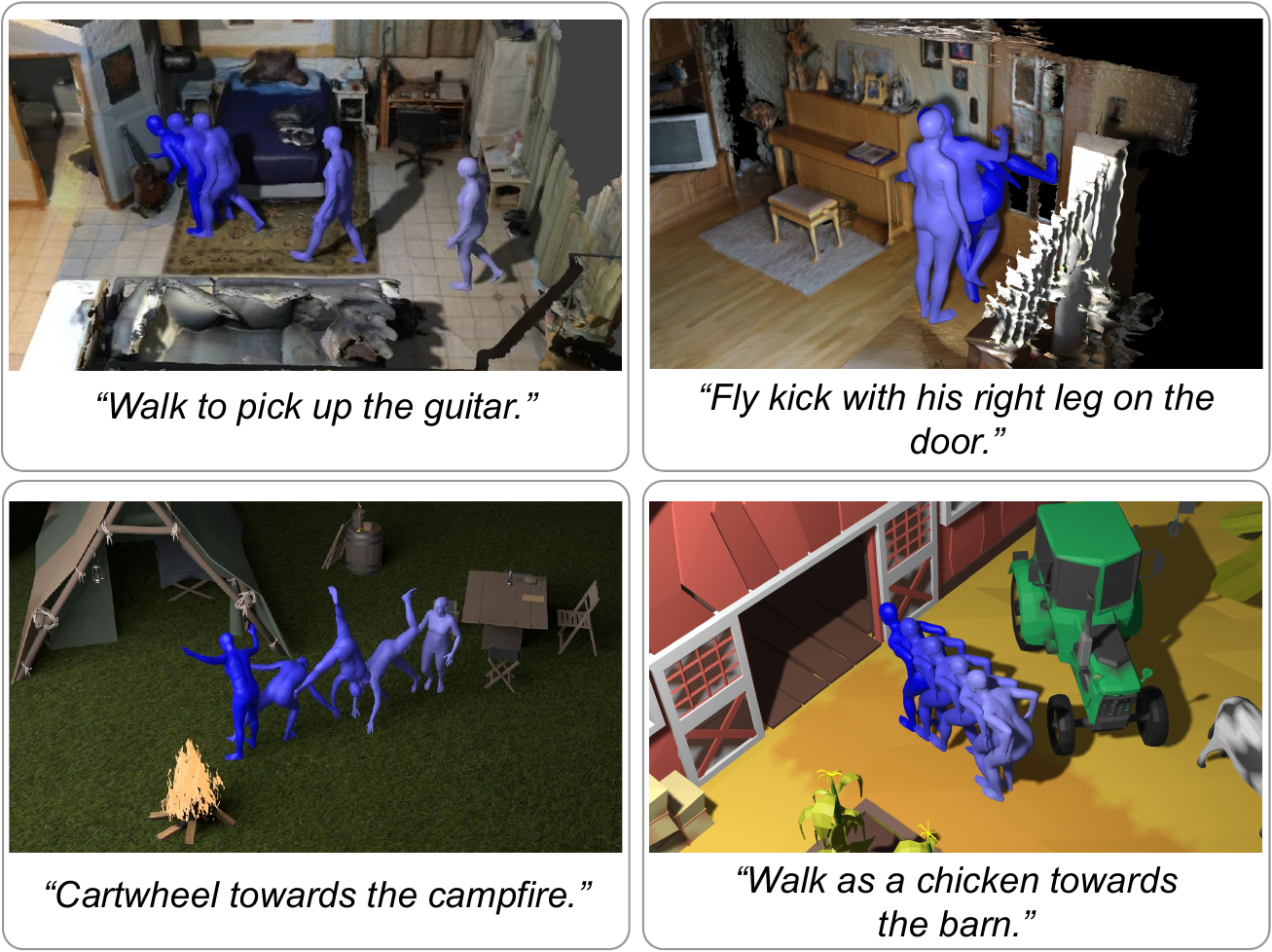}
  \caption{Illustration of scene-aware text-driven motion sequences generated by our TSTMotion framework in different 3D scenes based on text descriptions without any training. For clarity, as time progresses, human avatars in motion sequences transition from light to dark colors. More qualitative results in image and video formats are available in the supplementary.}
  \label{fig:illustration}
\end{figure} 

Text-to-motion generation, which aims to generate realistic human motion sequences based on the given text descriptions, has recently attracted significant research attention \cite{tevet2022human,zhang2022motiondiffuse}. Despite extensive efforts in this area, most research has focused solely on generating human motions with a blank background. However, humans often interact with different objects within diverse 3D scenes in real-world scenarios. For example, a human avatar in video games may be instructed to ``sit on the couch away from the TV''. To execute this instruction, the avatar needs to understand the motion semantic of the instruction (i.e., ``sit''), comprehend the context of the 3D scene (i.e., ``couch away from the TV''), and then perform the desired motion within the 3D scene \cite{wang2022humanise}. Consequently, traditional text-to-motion generators fail to meet the requirements of various real-world applications. To mitigate this challenge, it is crucial that the generated human motion sequences not only align with the given text descriptions but also appropriately interact with the given 3D scenes. Hence, scene-aware text-to-motion generation has become a valuable research area to advance various applications like game development, film creation, embodied AI, and virtual reality \cite{tevet2022human,xiao2024unified}.

To achieve scene-aware text-to-motion generation, one straightforward solution is to directly train a motion generator using scene-aware text-to-motion datasets \cite{wang2022humanise,jiang2024scaling,wang2024move}. However, this solution may be sub-optimal due to the limited scale and diversity of existing datasets. In particular, they lack a wide-ranging and varied assortment of combinations involving text, motion, and scene components. Take the HUMANISE dataset \cite{wang2022humanise} as an example: it is confined to indoor scenes and encompasses merely four actions: walking, sitting, lying, and standing. The challenges of scaling and diversifying such datasets are twofold: on one side, these datasets require numerous motion sequences interacting with a variety of objects and 3D environments, which even necessitates sophisticated motion capture systems \cite{Hassan_2021_ICCV}; on the other side, they require detailed annotations of object semantics and human poses, which involves significant manual labor. An alternative solution is to employ Reinforcement Learning (RL) for a human motion controller \cite{xiao2024unified,zhao2023synthesizing}. By leveraging the generalization capabilities of RL, this solution mitigates the reliance on the scale of datasets to some extent. However, this solution may necessitate more fine-grained datasets (e.g., objects annotated in part level) and face the intricacies associated with RL algorithms \cite{xiao2024unified}. 
Besides, existing methods often involve directly inputting point clouds into the model, which is inefficient due to the disordered and unstructured nature of point clouds.
In light of these challenges in scaling up training data, we aim to propose a training-free solution, thereby bypassing the need for collecting scene-aware text-to-motion datasets and training models with point clouds from scratch.

To realize a training-free approach for this task, it is promising to enable the scene awareness of existing blank-background text-to-motion generators, which have an extensive understanding of human motion. Nevertheless, this solution can present significant challenges: (1) These generators might struggle to comprehend 3D scenes, given that their original training input consists solely of text without scene information (e.g., ``sit down'') \cite{guo2022generating}. Consequently, devising an effective method to enable the generators’ scene awareness—particularly under a training-free paradigm—presents a considerable challenge. (2) Even if these generators can be scene-aware, it is still challenging to provide them with format-compatible and information-sufficient 3D scene knowledge in a training-free manner. To address these challenges, we introduce a novel framework, \textbf{TSTMotion} (\textbf{T}raining-free \textbf{S}cene-aware \textbf{T}ext-to-\textbf{Motion}), which pioneers a training-free method for scene-aware text-to-motion generation. 

Overall, TSTMotion aims to formulate a motion guidance integrated with text semantics and scene contexts, and then perform a training-free alignment between the motion guidance and blank-background motion generators.
To achieve such motion guidance, inspired by that foundation models contain extensive knowledge of human motions and 3D scenes, TSTMotion first incorporates three well-designed components based on foundation models, including: the Scene Compiler to represent the given 3D scene into a format comprehensible by the Motion Planner, the Motion Planner to offer a plausible motion guidance for the expected motion, and the Motion Checker to refine the generated motion guidance to ensure practical feasibility.
To effectively utilize the above-formulated motion guidance and achieve such alignment based on the motion guidance, TSTMotion further proposes to adapt existing blank-background motion generators (i.e., Motion Diffusion Models \cite{tevet2022human,zhang2022motiondiffuse}) through two training-free modifications: one ensures alignment of the generated motion sequences with the motion guidance; the other reduces the overlapping between motion sequences and 3D scenes. An overview of our proposed TSTMotion framework is illustrated in Fig.~\ref{fig:framework}.
Through these designs, TSTMotion refrains from being trained on specialized scene-aware text-to-motion datasets, and then exhibits increased diversity (i.e., not limited to common walking and sitting) and better generalization (i.e., indoor and outdoor 3D scenes) as shown in Fig.\ref{fig:illustration}. 

The contributions of our work are summarized as follows: 
1) We are the first to achieve scene-aware text-to-motion generation in a training-free manner, by our novel framework TSTMotion.
2) We incorporate Scene Compiler, Motion Planner and Motion Checker to craft and refine a motion guidance integrating text semantics with scene contexts, through which a blank-background motion generator with two modifications can generate scene-aware text-driven motion sequences in a training-free manner.
3) Extensive experiments demonstrate that TSTMotion achieves superior performance across various benchmarks without the need for specific scene-aware text-to-motion datasets or further training of any model. 
4) Furthermore, our framework exhibits the diversity of generated motions and generalizability to outdoor 3D scenes.

\begin{figure*}[t]
  \centering
  \includegraphics[width=0.9\linewidth,page=1]{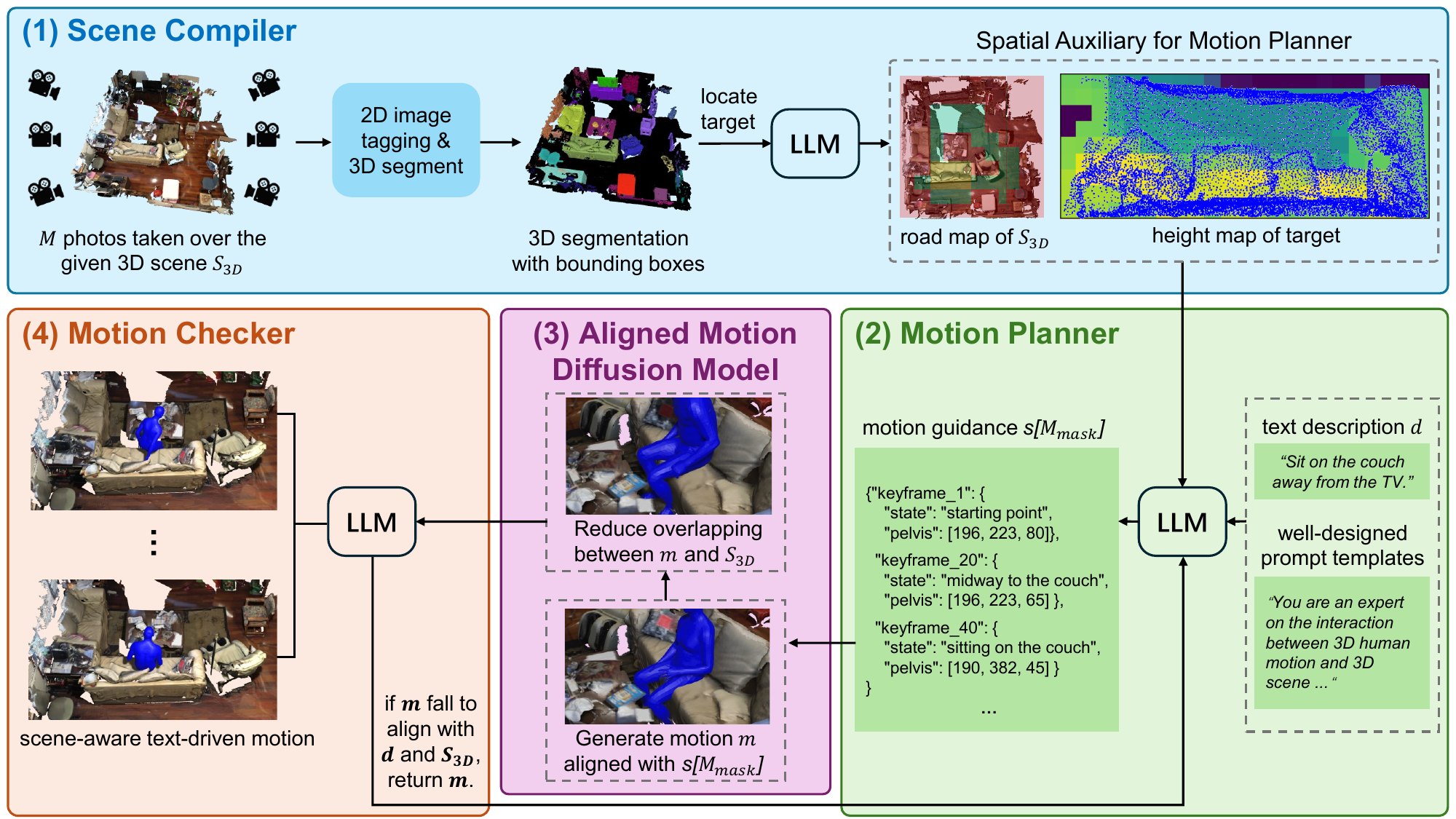}
  \caption{An overview of our proposed training-free TSTMotion framework for the given text $d$ and 3D scene $S_{3D}$. At first, the Scene Compiler extracts the spatial auxiliary in the $S_{3D}$. Based on the spatial auxiliary, the Motion Planner incorporates the text description and well-designed prompt templates to infer the motion guidance $s[M_{mask}]$. Equipped with the $s[M_{mask}]$, the Aligned Motion Diffusion Model predicts initial scene-aware text-driven motion sequences $m$ with two training-free modifications. Finally, the Motion Checker is applied to iteratively refine and generate the final $m$ to better align with the $d$ and $S_{3D}$.}
  \label{fig:framework}
\end{figure*} 

\section{METHODOLOGY}
In the context of training-free scene-aware text-to-motion generation, our goal is to generate realistic human motion sequences $m$ in a training-free manner, which not only aligns with the given text description $d$ (e.g., ``sit on the couch away from the TV''), but also properly interacts with the provided 3D scene $S_{3D}$. To this end, we propose a novel framework named TSTMotion as depicted in Fig.~\ref{fig:framework}, which sequentially incorporates four key components: Scene Compiler, Motion Planner, Aligned Motion Diffusion Mode,l and Motion Checker.

\subsection{Scene Compiler} \label{sec:3.1}
3D scenes are often represented as point clouds, posing a challenge to directly interpret. To address this issue, our Scene Compiler first translates the given 3D scene $S_{3D}$ into a spatial auxiliary that can be easily comprehended without training. When determining this spatial auxiliary, we recognize that the layout of 3D scenes and the shape of the target are more crucial for motion than their texture or appearance.
To this end, we propose that the Scene Compiler interprets the 3D scene $S_{3D}$ to LLMs as the road map of the $S_{3D}$ and the height map of the target. Specifically, the road map provides LLMs with walkable areas in the 3D scenes, while the height map provides the shape information of the target.

To interpret $S_{3D}$, we follow five steps: \textbf{(1) Recognition}: Identify object types in $S_{3D}$ by rendering it and capturing $M$ images from different angles. Use an image tagging model to identify objects and create a unified set of object vocabulary. \textbf{(2) Segmentation}: Use the object vocabulary and a 3D segmenter to determine the category and occupancy of each object in $S_{3D}$. \textbf{(3) Locating Target}: Simplify each object into a bounding box. Use an LLM to locate the target based on these bounding boxes. \textbf{(4) Road Map}: Project all bounding boxes onto the XOY Plane to create a road map indicating obstacles and the target. \textbf{(5) Height Map}: Select and project the point clouds of the target onto the XOY plane to form a height map, indicating the height of the point clouds.

\subsection{Motion Planner} \label{sec:3.2}
Leveraging the spatial auxiliary from the Scene Compiler, the Motion Planner aims to generate an understandable and desired motion guidance for the blank-background Motion Diffusion Models (MDMs) to interact with the scene.

Equipped with the spatial auxiliary, the primary consideration of the Motion Planner is to determine the format of the motion guidance. Typically, MDMs are trained on qualitative text descriptions such as ``walk forward'', which may not provide sufficient detail for accurate interaction with 3D scenes. Inspired by the idea that skeleton sequences $s$ (i.e., joint positions) serve as a simplified representation of motion sequences $m$ (i.e., joint rotations), we present motion guidance as skeleton sequences in 3D coordinates, denoted as $s \in R^{N \times J \times 3}$, where $N$ represents the number of frames and $J$ denotes the human body joints. Importantly, not all joints are relevant to both the text description and the 3D scene. Therefore, the skeleton sequence $s$ can only involve critical joints with a binary mask $M_{mask} \in \{0, 1\}^{N \times J}$. The resulting motion guidance can then be represented as $s[M_{mask}]$. 

So far, the input (i.e., spatial auxiliary) and output (i.e., motion guidance) of the LLM have been determined. Well-designed prompt templates are incorporated to leverage the reasoning capabilities of the LLM fully. Existing works \cite{wei2022chain} reveal that LLMs can better tackle a certain task if they can first divide the task into several simpler sub-tasks. Considering this, we also divide the task of the Motion Planner into several simpler sub-tasks (e.g., predict contact areas on the target), and guide the LLM to tackle these sub-tasks step-by-step. 

In summary, LLMs employed by the Motion Planner are capable of generating appropriate motion guidance in the form of $s[M_{mask}]$. This capability arises from two key factors: firstly, LLMs have demonstrated their ability to navigate on the map \cite{huang2023visual} and analyze the matrix data \cite{hegselmann2023tabllm}; secondly, LLMs have been pre-trained on an extensive corpus that includes rich descriptions of human skeletons and motions \cite{brown2020language}. 
Hence, LLMs contain implicit knowledge about the scene contexts and human behaviors, and can generate skeleton sequences plausibly interacting with scenes.

\subsection{Aligned Motion Diffusion Models} \label{sec:3.3}
We have now achieved an understandable motion guidance $s[M_{mask}]$ for the blank-background Motion Diffusion Models (MDMs). Inspired by posterior sampling \cite{chung2022diffusion}, we introduce two training-free modifications to empower the blank-background MDMs with scene awareness based on the motion guidance. The first modification aligns the generated motion sequences with the motion guidance, while the second modification reduces overlapping between motion sequences and 3D scenes. Importantly, \emph{these modifications are plug-and-play during inference, eliminating the necessity for additional training on specific datasets}.

\textbf{Revisiting motion diffusion models.} During the inference process, MDMs iteratively denoise noise $x_K \sim \mathcal{N}(\textbf{0}, \textbf{I})$ back into a clean motion $x_0$, formulated as: 
\begin{align}
&\epsilon_k \sim \mathcal{N}(0, \textbf{I}) \\
& \hat{x}_0^{k} = f_{MDM}(x_k, k, d) \\
& \mu_k = \frac{\sqrt{\overline{\alpha}_{k-1}}(1 - \alpha_k)}{1 - \overline{\alpha}_k} \hat{x}^{k}_0 + \frac{\sqrt{\alpha_k}(1-\overline{\alpha}_{k-1})}{1 - \overline{\alpha}_k} x_k \\
& x_{k-1} = \mu_k + \sqrt{(1-\alpha_k)} \epsilon_k \;
\end{align}
where $d$ denotes the given text prompt, $\{\alpha_k \in (0, 1)\}^K_{k=1}$ is a set of hyper-parameters, $\overline{\alpha}_k = \prod^k_{s=1} \alpha_s$, and $f_{MDM}$ denotes the MDM. Note that $\hat{x}^k_0$ represents the MDM's prediction of the final clean motion at step $k$.

\textbf{Modification 1: conditioning the MDM on motion guidance.} 
Here, we aim to condition the MDM $f_{MDM}$ on the motion guidance $s[M_{mask}]$, thereby aligning the generated motion sequence $m$ (i.e., $x_0$) with both the text description $d$ and the 3D scene $S_{3D}$. To achieve this, we recall that $\hat{x}^k_0$ represents the MDM's prediction of the final clean motion at step $k$. Therefore, it is potential to utilize the alignments between $s[M_{mask}]$ and $\hat{x}^k_0$, introducing $L_{align}$ for the MDM:
\begin{align}
& L_{align}(s[M_{mask}], \hat{x}^k_0) = ||s[M_{mask}] - FK(\hat{x}^k_0)[M_{mask}]||_2^2
\end{align}
where $FK$ is a forward kinematic function to map motion sequences $\hat{x}^k_0$ (i.e., joint rotations) into the skeleton sequence $s$ (i.e., joint positions), $L_{align}$ is a function to measure the distance between $\hat{x}^k_0$ and $s[M_{mask}]$.

Then, we propose an effective yet straightforward modification, namely reducing the gap between $s[M_{mask}]$ and $\hat{x}^k_0$ by adding the gradient of $L_{align}$ on $\hat{x}^k_0$. Namely, in each step, the motion prediction $\hat{x}^k_0$ is modified as:
\begin{align} \label{eq:diffusion_6}
& \hat{x}^k_0 \leftarrow \hat{x}^k_0 - \lambda\cdot\nabla_{x_k} L_{align}(s[M_{mask}], \hat{x}^k_0)
\end{align}
where $\lambda$ is a hyper-parameter to control strength. Such modification narrows down the output distribution towards the specific distribution that accurately matches the 
$s[M_{mask}]$.

\textbf{Modification 2: reducing overlapping between motion sequences and 3D scene.} 
Ideally, the generated motion sequences have aligned with both the given text description and the 3D scene through motion guidance. However, motion sequences may overlap with the 3D scene, as the motion guidance can not be flawless. To address this, we introduce an additional modification to reduce the overlapping between the generated motion sequences and the 3D scene.

Through the motion guidance, the semantics of the generated motion sequences are already aligned with the semantics of the text. Thus, the primary consideration is to prevent motion sequences from overlapping with the scene while preserving their motion semantics, which only necessitates minor adjustments to motion sequences. To this end, we incorporate the Signed Distance Field (SDF) function to evaluate the distance from the points of motion’s mesh to the scene’s mesh. Then, an object $L_{scene}$ is proposed to evaluate how deep the points inside the scene's mesh:
\begin{align}
& L_{scene}(\hat{x}^k_0,P) = ReLU(-SDF(SMPL(\hat{x}^k_0), P))
\end{align}
where $P$ denotes the mesh of the given scene. The function $SMPL$ is used to skin the skeleton with the widely used SMPL model \cite{SMPL-X:2019}. The function $SDF$ is used to query the signed distance of the points of SMPL compared to the scene' mesh $P$, and $SDF$ will be updated in each reverse process of MDM. The function $ReLU$ is used to exclude points of SMPL that are not inside the scene' mesh. 

Subsequently, we are able to penalize and reduce the overlapping between motion sequences and 3D scenes, by adding the gradient of $L_{scene}$ on $\hat{x}^k_0$:
\begin{align}
& \hat{x}^k_0 \leftarrow \hat{x}^k_0 - \eta\cdot\nabla_{x_k} L_{scene}(\hat{x}^k_0,P)  \label{eq:scene}
\end{align}
where $\eta$ is the hyper-parameter to control strength.

Notably, the functions mentioned above are all differentiable and the entire sampling process of MDM does not require training any model with ground-truth data.

\subsection{Motion Checker} \label{sec:3.4}
In order to improve the robustness of TSTMotion, we propose a Motion Checker to further align the generated motion sequences with the text description and the 3D scene. Specifically, the Motion Checker, utilizing an LLM with well-designed prompt templates, is responsible for verifying whether the generated motion sequences meet the task requirements. For instance, it verifies whether the motion is beyond the scene and matches the motion semantics. If the generated motion sequences fail to meet these criteria, the framework initiates a restart from the Motion Planner. This iterative process is feasible because the Motion Checker focuses on identifying and rectifying potential errors rather than generating motion from scratch \cite{zhang2024enhancing}. Notably, only one iteration is necessary to make TSTMotion achieve superior performance.

\begin{figure*}[!th]
  \centering
    \includegraphics[width=0.9\linewidth]{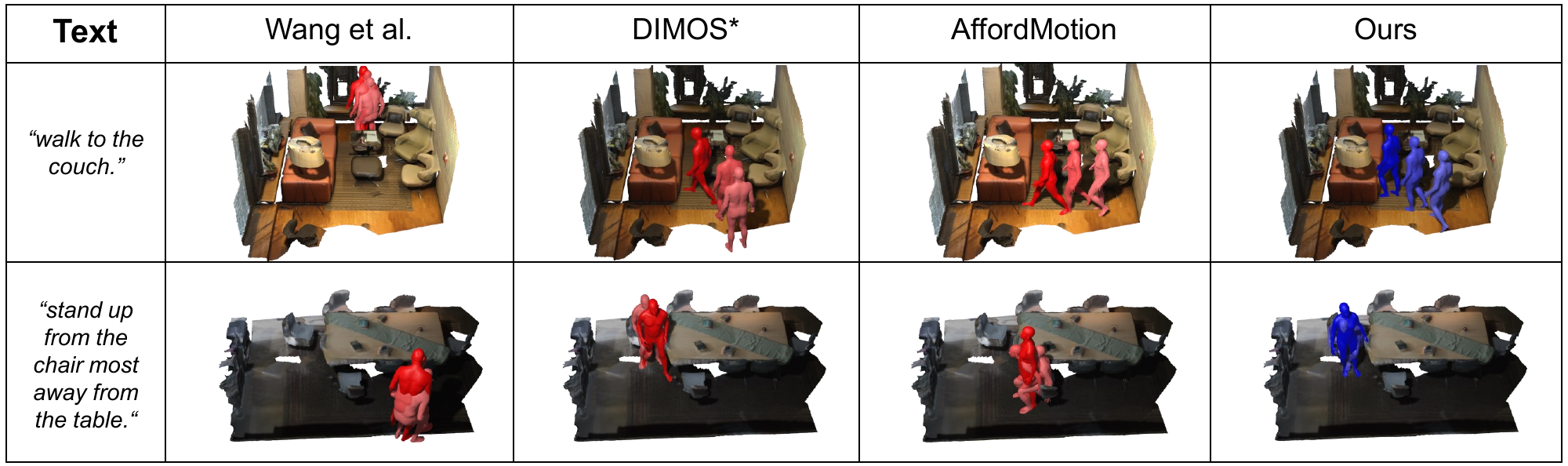}
   \caption{Comparison of between Wang et al.\cite{wang2022humanise}, DIMOS*, AffordMotion and our TSTMotion on the unseen PROX dataset.}
   \label{fig:supp_PROX}
\end{figure*}

\section{Experiments}

\begin{table}[!t]
\centering
\caption{Results on the testing dataset of the HUMANISE.}
\resizebox{\linewidth}{!}{
\begin{tabular}{c|c|c|c|c|c} \hline
Method & quality score \textuparrow & action score \textuparrow & body-to-goal distance \textdownarrow & non-collision score \textuparrow & contact score \textuparrow \\ \hline
Wang et al. & 2.74$^{\pm1.12}$ & 3.67$^{\pm0.59}$ & 1.01 & 0.9988 & 0.73 \\
DIMOS* & 2.87$^{\pm0.55}$ & 3.79$^{\pm0.21}$ & 0.73 & \textbf{0.9991} & 0.86 \\
Ours & \textbf{2.94}$\mathbf{^{\pm0.86}}$ & \textbf{3.81}$\mathbf{^{\pm0.37}}$ & \textbf{0.71} & \textbf{0.9991} & \textbf{0.91} \\ \hline
\end{tabular}
}
\label{Tab:HUMANISE}
\end{table}

\begin{table*}[!t]
\caption{Results on the novel testing dataset from AffordMotion.}
\resizebox{\linewidth}{!}{
\small
\centering
\begin{tabular}{c|c|c|c|c|c|c|c|c|c} \hline
Method & R-Preci.\textuparrow & FID\textdownarrow & Multi. Dist.\textdownarrow & Diversity$\rightarrow$ & MultiModality\textuparrow & \makecell{contact \\ score \textuparrow} & \makecell{non-coll. \\ score \textuparrow} & \makecell{quality \\ score \textuparrow} & \makecell{action \\ score \textuparrow} \\ \hline
Grountruth & 0.875$^{\pm.002}$ & 0.000$^{\pm.000}$ & 3.342$^{\pm.004}$ & 9.442$^{\pm.301}$ & - & - & -  & -  & - \\ \hline
AffordMotion & 0.478$^{\pm.069}$ & 7.887$^{\pm1.19}$ & 6.226$^{\pm.261}$ & \textbf{7.935}$\mathbf{^{\pm.857}}$ & \textbf{5.15}9$\mathbf{^{\pm.356}}$ & 71.98$^{\pm2.54}$ & 99.83$^{\pm.006}$ & 2.39$^{\pm1.09}$ & 2.85$^{\pm1.77}$ \\ 
Ours  & \textbf{0.592}$\mathbf{^{\pm.047}}$ & \textbf{6.739}$\mathbf{^{\pm.082}}$ & \textbf{5.513}$\mathbf{^{\pm.327}}$ & 5.724$^{\pm.820}$ & 4.713$^{\pm.620}$ & \textbf{86.12}$\mathbf{^{\pm1.33}}$ & \textbf{99.88}$\mathbf{^{\pm.005}}$  & \textbf{2.55}$\mathbf{^{\pm1.33}}$ & \textbf{3.07}$\mathbf{^{\pm1.36}}$ \\ 
\hline
\end{tabular}
}
\label{Tab:AffordMotion}
\end{table*}

To evaluate the efficacy of our proposed TSTMotion, in this section, we conduct both quantitative and qualitative evaluations. For quantitative evaluations, we compare TSTMotion on the testing dataset of HUMANISE \cite{wang2022humanise} AffordMotion \cite{wang2024move} and DIMOS \cite{zhao2023synthesizing}. For qualitative evaluations, we also compare TSTMotion with these methods on several indoor and outdoor 3D scenes. Our comprehensive evaluation demonstrates the superiority of TSTMotion in generating scene-aware text-driven motion sequences, and generalizing to novel, diverse, and challenging scenarios.

\subsection{Datasets and Evaluation Metrics}

\textbf{Datasets.} We evaluate our framework using the following datasets. Following the evaluation setup in Wang et al. \cite{wang2022humanise}, we here evaluate our framework on the HUMANISE testing dataset, which contains 19.6k human motion sequences in 643 different 3D scenes. Moreover, we further evaluate our framework on the novel testing dataset from AffordMotion \cite{wang2024move}, which comprises 16 scenes from diverse sources along with 80 crafted descriptions.

\textbf{Evaluation metrics.} To evaluate the quality of the motion generated in different 3D scenes, we incorporate several metrics to evaluate whether the generated motion sequences are consistent with the given text and properly interact with the given 3D scene.
Following Wang et al. \cite{wang2022humanise}, we use the following three metrics: the quality score and the action score to measure the overall quality and the action-semantic accuracy of the generated motion sequences by the perceptual study; the contact score to evaluate whether the distance between the body and the scene is under a pre-defined threshold; the non-collision score to evaluate the non-overlapping degree between the human and objects in scenes; the body-to-goal distance to calculate the shortest distance (in meters) between the target object and the human motion. For quality and action scores, mean and standard deviation are reported.
Following Wang et al. \cite{wang2024move}, we adopt five additional metrics: MultiModality to evaluate the variation compared to text descriptions; R-Precision and Multimodal Distance to assess the correlation between generated motions and the given text; FID to evaluate the difference between the distributions of generated motion sequences and ground truth. All evaluations are conducted five times to ensure robustness, with a 95\% confidence interval indicated by $\pm$.

\begin{table}[!t]
\small
\caption{Evaluation on the design choices incorporated in the TSTMotion on the testing dataset of HUMANISE.}
\resizebox{\linewidth}{!}{
\centering
\begin{tabular}{c|c|c|c} \hline
Method  & body-to-goal \textdownarrow & non-collision \textuparrow & contact \textuparrow \\ \hline
baseline & 0.95 & 0.9975 & 0.79 \\
w/o spatial auxiliary & 0.72 & 0.9977 & 0.90 \\
w/o modification 1 & 0.93 & 0.9982 & 0.84 \\
w/o modification 2 & 0.74 & 0.9974 & 0.81  \\
w/o Motion Checker & 0.72 & 0.9989 & 0.89  \\
Ours & \textbf{0.71} & \textbf{0.9991} & \textbf{0.91} \\ 
\hline
\end{tabular}
}
\label{Tab:ablation_study_1}
\end{table}

\subsection{Implementation Details}
Our experiments are conducted on one RTX 3090 GPU. For the 2D image tagging and 3D segmenter, we use RAM \cite{zhang2024recognize} and OpenIns3D \cite{huang2023openins3d} respectively. For the deployed LLMs, we all use the GPT-4 \cite{achiam2023gpt}. For the motion diffusion model, we use the Xie et al. \cite{xie2023omnicontrol} Besides, we set the coefficient hyper-parameter $\lambda$ used in Eq.~\ref{eq:diffusion_6} to be 2, set hyper-parameter $\eta$ used in Eq.~\ref{eq:scene} to be 0.5, set the number of photos $M$ to be 16 following Huang et al. \cite{huang2023openins3d}, and set the number of iterations for Motion Checker to be 1.

\subsection{Main Results}
\textbf{HUMANISE.} As demonstrated in Table~\ref{Tab:HUMANISE}, TSTMotion exhibits superior performance across a range of evaluation metrics on the comprehensive testing dataset of HUMANISE. Notably, DIMOS cannot locate objects in the scene by itself. Therefore, we introduce our Scene Compiler to assist DIMOS in locating objects denoted as DIMOS*. The lower body-to-goal distances of TSTMotion and DIMOS* further validates the localization capability of the Scene Compiler.

\textbf{AffordMotion.} Evaluating our framework on the novel testing dataset from AffordMotion reveals that our training-free framework yields competitive outcomes across various metrics, as illustrated in Table \ref{Tab:AffordMotion}. Specifically, our framework not only retains sufficient semantics (i.e., higher R-Precision and lower MultiModal Distance), but also matches the scene appropriately (i.e., higher contact and non-collision score).

\textbf{Qualitative results.} In addition to quantitative comparisons, we showcase the qualitative process of our framework within different settings. As shown in Fig.~\ref{fig:supp_PROX}, in the PROX dataset \cite{PROX:2019}, Wang et al.\cite{wang2022humanise} fails to generate plausible motion sequences. Such poor performance may be caused by the limited scale and diversity of training dataset. Although AffordMotion can generate reasonable motions, it can suffer from localization in complex sentences. Meanwhile, our TSTMotion can generate more realistic motion sequences, which further validate the capability of TSTMotion. As shown in Figures \ref{fig:illustration}, it highlights the versatility of TSTMotion in generating diverse motion sequences such as cartwheeling. On the contrary, existing methods typically concentrate on a limited set of actions such as walking, sitting, standing, and lying. Further, our framework demonstrates its adaptability and robustness across unseen outdoor 3D scenes, thanks to that TSTMotion leverages established foundation models and is not specialized for any specific dataset focusing on the indoor scenes.

\subsection{Ablation Studies}
Following Wang et al. \cite{wang2024move}, we further examine the impact of different settings of TSTMotion on the testing dataset of HUMANISE with three objective metrics. \textbf{More ablation studies are available in the supplementary.}

\textbf{Impact of the design choices in the TSTMotion.} To evaluate the efficacy of design choices in TSTMotion, we test several variants of TSTMotion: 
In the first variant (baseline), we directly condition the reverse diffusion process on $s[M_{mask}]$ without the spatial auxiliary, the two modifications of Aligned MDM and the Motion Checker. 
In the second variant (w/o spatial auxiliary), we directly provide the bounding boxes instead of incorporating the spatial auxiliary.
In the third variant (w/o modification 1), we perform the alignment by directly inputting $s[M_{mask}]$ into the Aligned MDM, instead of aligning $\hat{x}_0^k$ with $s[M_{mask}]$. 
In the fourth variant (w/o modification 2), we do not incorporate modification 2 to reduce overlapping.
In the fifth variant (w/o Motion Checker), we do not incorporate the Motion Checker and the iteration process. 
As shown in Table \ref{Tab:HUMANISE}, our framework outperforms all variants. This shows the effectiveness of the modifications for the aligned MDM, the spatial auxiliary to explicitly boost the LLM's reasoning capability and the Motion Checker to iteratively refine the generated motion. 

\section{Conclusion}
In this paper, we have proposed a novel scene-aware text-to-motion generation framework TSTMotion. Specifically, we design a motion guidance that can be crafted and refined by Scene Compiler, Motion Planner and Motion Checker. By utilizing the motion guidance, a Motion Diffusion Model with two training-free modifications can generate scene-aware text-driven motion sequences. Without requiring any specific scene-aware text-to-motion datasets or further model training, our framework generalizes well to different indoor and outdoor 3D scenes, and achieves a superior result compared to the previous training-based method.

\bibliographystyle{IEEEbib}
\bibliography{icme2025references}

\begin{thebibliography}{10}

\bibitem{tevet2022human}
Guy Tevet, Sigal Raab, Brian Gordon, Yonatan Shafir, Daniel Cohen-Or, and Amit~H Bermano,
\newblock ``Human motion diffusion model,''
\newblock {\em arXiv preprint arXiv:2209.14916}, 2022.

\bibitem{zhang2022motiondiffuse}
Mingyuan Zhang, Zhongang Cai, Liang Pan, Fangzhou Hong, Xinying Guo, Lei Yang, and Ziwei Liu,
\newblock ``Motiondiffuse: Text-driven human motion generation with diffusion model,''
\newblock {\em arXiv preprint arXiv:2208.15001}, 2022.

\bibitem{wang2022humanise}
Zan Wang, Yixin Chen, Tengyu Liu, Yixin Zhu, Wei Liang, and Siyuan Huang,
\newblock ``Humanise: Language-conditioned human motion generation in 3d scenes,''
\newblock {\em NeurIPS}, vol. 35, pp. 14959--14971, 2022.

\bibitem{xiao2024unified}
Zeqi Xiao, Tai Wang, Jingbo Wang, Jinkun Cao, Wenwei Zhang, Bo~Dai, Dahua Lin, and Jiangmiao Pang,
\newblock ``Unified human-scene interaction via prompted chain-of-contacts,''
\newblock in {\em ICLR}, 2024.

\bibitem{jiang2024scaling}
Nan Jiang, Zhiyuan Zhang, Hongjie Li, Xiaoxuan Ma, Zan Wang, Yixin Chen, Tengyu Liu, Yixin Zhu, and Siyuan Huang,
\newblock ``Scaling up dynamic human-scene interaction modeling,''
\newblock in {\em CVPR}, 2024, pp. 1737--1747.

\bibitem{wang2024move}
Zan Wang, Yixin Chen, Baoxiong Jia, Puhao Li, Jinlu Zhang, Jingze Zhang, Tengyu Liu, Yixin Zhu, Wei Liang, and Siyuan Huang,
\newblock ``Move as you say, interact as you can: Language-guided human motion generation with scene affordance,''
\newblock {\em arXiv preprint arXiv:2403.18036}, 2024.

\bibitem{Hassan_2021_ICCV}
Mohamed Hassan, Duygu Ceylan, Ruben Villegas, Jun Saito, Jimei Yang, Yi~Zhou, and Michael~J. Black,
\newblock ``Stochastic scene-aware motion prediction,''
\newblock in {\em ICCV}, October 2021, pp. 11374--11384.

\bibitem{zhao2023synthesizing}
Kaifeng Zhao, Yan Zhang, Shaofei Wang, Thabo Beeler, and Siyu Tang,
\newblock ``Synthesizing diverse human motions in 3d indoor scenes,''
\newblock {\em arXiv preprint arXiv:2305.12411}, 2023.

\bibitem{guo2022generating}
Chuan Guo, Shihao Zou, Xinxin Zuo, Sen Wang, Wei Ji, Xingyu Li, and Li~Cheng,
\newblock ``Generating diverse and natural 3d human motions from text,''
\newblock in {\em CVPR}, 2022, pp. 5152--5161.

\bibitem{wei2022chain}
Jason Wei, Xuezhi Wang, Dale Schuurmans, Maarten Bosma, Fei Xia, Ed~Chi, Quoc~V Le, Denny Zhou, et~al.,
\newblock ``Chain-of-thought prompting elicits reasoning in large language models,''
\newblock {\em NeurIPS}, vol. 35, pp. 24824--24837, 2022.

\bibitem{huang2023visual}
Chenguang Huang, Oier Mees, Andy Zeng, and Wolfram Burgard,
\newblock ``Visual language maps for robot navigation,''
\newblock in {\em ICRA}. IEEE, 2023, pp. 10608--10615.

\bibitem{hegselmann2023tabllm}
Stefan Hegselmann, Alejandro Buendia, Hunter Lang, Monica Agrawal, Xiaoyi Jiang, and David Sontag,
\newblock ``Tabllm: Few-shot classification of tabular data with large language models,''
\newblock in {\em International Conference on Artificial Intelligence and Statistics}. PMLR, 2023, pp. 5549--5581.

\bibitem{brown2020language}
Tom Brown, Benjamin Mann, Nick Ryder, Melanie Subbiah, Jared~D Kaplan, Prafulla Dhariwal, Arvind Neelakantan, Pranav Shyam, Girish Sastry, Amanda Askell, et~al.,
\newblock ``Language models are few-shot learners,''
\newblock {\em NeurIPS}, vol. 33, pp. 1877--1901, 2020.

\bibitem{chung2022diffusion}
Hyungjin Chung, Jeongsol Kim, Michael~T Mccann, Marc~L Klasky, and Jong~Chul Ye,
\newblock ``Diffusion posterior sampling for general noisy inverse problems,''
\newblock {\em arXiv preprint arXiv:2209.14687}, 2022.

\bibitem{SMPL-X:2019}
Georgios Pavlakos, Vasileios Choutas, Nima Ghorbani, Timo Bolkart, Ahmed A.~A. Osman, Dimitrios Tzionas, and Michael~J. Black,
\newblock ``Expressive body capture: {3D} hands, face, and body from a single image,''
\newblock in {\em CVPR}, 2019, pp. 10975--10985.

\bibitem{zhang2024enhancing}
Hang Zhang, Wenxiao Zhang, Haoxuan Qu, and Jun Liu,
\newblock ``Enhancing human-centered dynamic scene understanding via multiple llms collaborated reasoning,''
\newblock {\em arXiv preprint arXiv:2403.10107}, 2024.

\bibitem{zhang2024recognize}
Youcai Zhang, Xinyu Huang, Jinyu Ma, Zhaoyang Li, Zhaochuan Luo, Yanchun Xie, Yuzhuo Qin, Tong Luo, Yaqian Li, Shilong Liu, et~al.,
\newblock ``Recognize anything: A strong image tagging model,''
\newblock in {\em CVPR}, 2024, pp. 1724--1732.

\bibitem{huang2023openins3d}
Zhening Huang, Xiaoyang Wu, Xi~Chen, Hengshuang Zhao, Lei Zhu, and Joan Lasenby,
\newblock ``Openins3d: Snap and lookup for 3d open-vocabulary instance segmentation,''
\newblock {\em arXiv preprint arXiv:2309.00616}, 2023.

\bibitem{achiam2023gpt}
Josh Achiam, Steven Adler, Sandhini Agarwal, Lama Ahmad, Ilge Akkaya, Florencia~Leoni Aleman, Diogo Almeida, Janko Altenschmidt, Sam Altman, Shyamal Anadkat, et~al.,
\newblock ``Gpt-4 technical report,''
\newblock {\em arXiv preprint arXiv:2303.08774}, 2023.

\bibitem{xie2023omnicontrol}
Yiming Xie, Varun Jampani, Lei Zhong, Deqing Sun, and Huaizu Jiang,
\newblock ``Omnicontrol: Control any joint at any time for human motion generation,''
\newblock {\em arXiv preprint arXiv:2310.08580}, 2023.

\bibitem{PROX:2019}
Mohamed Hassan, Vasileios Choutas, Dimitrios Tzionas, and Michael~J. Black,
\newblock ``Resolving {3D} human pose ambiguities with {3D} scene constraints,''
\newblock in {\em International Conference on Computer Vision}, Oct. 2019, pp. 2282--2292.

\end{thebibliography}

\end{document}